\documentclass{article}

\usepackage[utf8]{inputenc} % allow utf-8 input
\usepackage{hyperref}      % hyperlinks
\usepackage{url}            % simple URL typesetting
\usepackage{booktabs}       % professional-quality tables
\usepackage{amsfonts}       % blackboard math symbols
\usepackage{nicefrac}       % compact symbols for 1/2, etc.
\usepackage{microtype}      % microtypography
\usepackage{color}                %AMS Packages
\usepackage{amssymb}
\usepackage{amsmath}
\usepackage{amsthm}
\usepackage{graphicx}
\usepackage[authoryear]{natbib}
    \usepackage{algorithmicx}
    \usepackage[ruled]{algorithm}
    \usepackage{algpseudocode}

\newcommand{\abs}[1]{\ensuremath{|{#1}|}}

\DeclareMathOperator*{\argmax}{argmax}

\newcommand*{\argmaxl}{\argmax\limits}

\begin{document}
% \nipsfinalcopy is no longer used

\begin{center}
	{\centering{\LARGE\bf\it Feature selection in functional data classification with recursive maxima hunting}}\begin{center}
		
	\end{center}
\end{center}%, From big data to {\it data learning}, {\it Data learning } through big data} %\LaTeX\ template\tnoteref{mytitlenote}
%\tnotetext[mytitlenote]{Fully documented templates are available in the elsarticle package on \href{http://www.ctan.org/tex-archive/macros/latex/contrib/elsarticle}{CTAN}.}

%% Group authors per affiliation:\begin{center}

\begin{center}
	{\Large Jos\'e L. Torrecilla} \\
	Computer Science Department. Universidad Aut\'onoma de Madrid, Spain\\joseluis.torrecilla@uam.es\\ \vspace{20pt}
	{\Large  Alberto Su\'arez}\\
	Computer Science Department. Universidad Aut\'onoma de Madrid, Spain
\end{center}

\begin{abstract}
Dimensionality reduction is one of the key issues in the design of 
effective machine learning methods for automatic induction.  
In this work, we introduce recursive maxima hunting (RMH) for variable
selection in classification problems with functional data.
In this context, variable selection techniques are especially attractive 
because they reduce the dimensionality, facilitate the interpretation 
and can improve the accuracy of the predictive models.
The method, which is a recursive extension of 
maxima hunting (MH), performs variable selection by identifying the maxima 
of a relevance function, which measures the strength of the correlation of the 
predictor functional variable with the class label. 
At each stage, the information associated 
with the selected variable is removed by subtracting
the conditional expectation of the process.
The results of an extensive empirical evaluation are used to 
illustrate that, in the problems investigated, RMH has comparable or 
higher predictive accuracy than standard dimensionality reduction
techniques, such as PCA and PLS, and state-of-the-art
feature selection methods for functional data, such as maxima hunting.
\end{abstract}

\section{Introduction}

In many important prediction problems from 
different areas of application (medicine, environmental monitoring, etc.)
the data are characterized by a function, instead of by a vector of attributes, 
as is commonly assumed in standard machine learning problems.
Some examples of these types of data are functional magnetic 
resonance imaging (fMRI) \citep{gro08} and near-infrared spectra (NIR) \citep{xia10}. 
Therefore, it is important to develop methods for automatic induction 
that take into account the functional structure of the data (infinite dimension, 
high redundancy, etc.) \citep{ram05,fer06}. 
In this work, the problem of classification of functional data is addressed.
For simplicity, we focus on binary classification problems \citep{bai11}. 
Nonetheless, the proposed method can be readily extended to a multiclass setting. 
Let $X(t), t\in[0,1]$ be a continuous stochastic process in a probability space
 $(\Omega,\mathcal{F},\mathbb{P})$. 
 A functional datum $X_n(t)$ is a realization of this process (a trajectory). 
Let $\{X_n(t), Y_n\}_{n=1}^{N_{train}}, t \in [0,1]$ be a set of trajectories
labeled by the dichotomous variable $Y_n \in \{0,1\}$. These trajectories come   
from one of two different populations; 
either $P_0$, when the label is $Y_n = 0$, or  $P_1$, when the label is $Y_n = 1$.
For instance, the data could be the ECG's from either healthy or sick persons ($P_0$ and $P_1$, respectively).
The classification problem consist in deciding to which population a new 
unlabeled observation $X^{test}(t)$ belongs 
(e.g., to decide from his or her ECG whether a person is healthy or not). 
Specifically, we are interested in the problem of dimensionality 
reduction for functional data classification. The goal is to achieve the optimal 
discrimination performance using only a finite, 
small set of values from the trajectory as input to a standard classifier 
(in our work, $k$-nearest neighbors).

In general, to properly handle functional data, some kind of reduction of information is
necessary. Standard dimensionality reduction methods in functional data analysis (FDA) are based on principal component analysis (PCA) \mbox{\citep{ram05}} 
or partial least squares (PLS)  \mbox{\citep{pre07}}.
In this work, we adopt a different approach based on variable selection \citep{guy06}. 
The goal is to replace the complete function $X(t)$ by a $d$-dimensional vector ($X(t_1),\ldots,X(t_d)$) 
for a set of ``suitable chosen'' points $\{t_1,\ldots,t_d\}$ (for instance, instants in a heartbeat in ECG's), 
where $d$  is small.
% The application of feature selection techniques in the functional context 
% is theoretically supported in some non-trivial models 
% for which the optimal classification rule depends on $X(t)$ 
% through the values of the process only at a finite subset of points \citep{ber16mh,ber16rkhs}.
% As reported in the existing bibliography and illustrated in our experiments,
% feature selection can improve the accuracy of standard classifiers
% in many functional classification problems of practical interest, in which 
% the optimal rule is unknown \citep{del12cw,ber16mh}. 
% Furthermore, since the reduction is made in terms of the original variables, 
% feature selection techniques renders the classification process more understandable. 
% This is an advantage with respect to other information reduction
% methodologies such as PLS or PCA, 
% in which the selected components are difficult to interpret because they 
% involve the complete trajectory.  

Most previous work on feature selection in supervised learning with functional data 
is quite recent and focuses on regression problems; for instance, on the analysis of
fMRI images \citep{gro08,rya10} and NIR spectra \citep{xia10}. In particular, 
adaptations of {\it lasso} and other embedded methods have been proposed 
to this end (see, e.g., \cite{kne11,zho13,ane14}).  
In most cases, functional data are simply treated as high-dimensional vectors 
for which the standard methods apply. 
Specifically, \cite{gom09} propose  feature extraction from the functional trajectories 
before applying a multivariate variable selector based on measuring the mutual information. Similarly, \cite{fer15} compare  different 
standard feature selection techniques for image texture classification. 
The method of  minimum Redundancy Maximum Relevance (mRMR) 
introduced by \cite{din05} has been
applied to functional data in \cite{ber16mrmr}.
In that work distance correlation \citep{sze07} is used instead of mutual information
to measure nonlinear dependencies, with good results. 
A fully functional perspective is adopted in 
\cite{fer10} and \cite{del12cw}. 
In these articles, a wrapper approach is used to select 
the optimal set of instants in which the trajectories should 
be monitored by minimizing a cross-validation estimate of the classification error. 
\cite{ber16rkhs} introduce a filter selection procedure based on computing the 
Mahalanobis distance and Reproducing Kernel Hilbert Space techniques. 
Logistic regression models have been
 applied to the problem of binary classification with functional data in 
\cite{lin09} and \cite{mck10}, assuming Brownian and fractional Brownian trajectories, respectively. 
Finally, the selection of intervals or elementary functions  
instead of variables is addressed in \cite{li08, fra16} or \cite{tia13}.
% Even though wrapper algorithms have the potential of reaching higher accuracy, they are
% generally more costly computationally than filter methods. Furthermore, 
% the interpretation of the predictive model is less clear because 
% the feature selection process is classifier-dependent.
% In this work we will focus on filter methods, which are generally faster, 
% ``model-free'' and have also good predictive accuracy in practice.

From the analysis of previous work one concludes that, in general, it is 
preferable, both in terms of accuracy and interpretability, 
to adopt a fully functional approach to the problem. 
In particular, if the data are characterized by functions that are continuous,
values of the trajectory that are close to each other tend to be highly redundant
and convey similar information. Therefore, if the value of the process at 
a particular instant has high discriminant capacity, one could think of
discarding nearby values. This idea is exploited in maxima hunting (MH) \citep{ber16mh}.
% In MH one selects the local maxima of a relevance function that measures 
% the dependence between the $X(t)$  and the class label. 
% MH is fully functional, has good empirical performance, and 
% its results are easy to interpret. However, it also presents some drawbacks: 
% examples can be found in which some relevant points for classification are 
% not local maxima of the relevance function. Furthermore, the local maxima 
% are not necessarily independent of each other. 

In this work, we introduce recursive Maxima Hunting (RMH), a novel variable selection method 
for feature selection in functional data classification that takes advantage of the good properties of MH while addressing some of its deficiencies. 
The extension of MH consists in removing the information conveyed by each selected local maximum before
searching for the next one in a recursive manner. 
The rest of the paper is organized as follows:
Maxima hunting for feature selection in classification problems 
with functional data is introduced in Section \ref{sec:mh}. 
Recursive maxima hunting, which is the method proposed in this work, is described 
in  Section \ref{sec:mhrec}.  The improvements that can be obtained with this novel 
feature selection method are analyzed in an exhaustive empirical evaluations whose 
results are presented and discussed in Section \ref{sec:exp}.

\section{Maxima Hunting}\label{sec:mh}
Maxima hunting (MH) is a method for feature selection in functional classification 
based on measuring dependencies between values selected from  
$\left\{ X(t), t \in [0,1] \right\}$ and the response variable \citep{ber16mh}.
In particular, one selects the values $\left\{X(t_1), \ldots, X(t_d) \right\}$ 
whose dependence with the class label (i.e., the response variable) 
is locally maximal. 
 Different measures of dependency can be used for this purpose. 
In \cite{ber16mh}, the authors propose the distance correlation \citep{sze07}. 
The distance covariance between the random variables 
$X \in {\mathbb R}^p$ and $Y \in {\mathbb R}^q$, whose components are assumed to
have finite first-order moments, is 
\begin{equation}
{\cal V}^2(X,Y) = \int_ {\mathbb{R}^{p+q}} \mid \varphi_{X,Y}(u,v) -\varphi_X(u) \varphi_Y(v)\mid^2 w(u,v) du dv,
\end{equation}
where $\varphi_{X,Y}$, $\varphi_{X}$, $\varphi_{Y}$ are the characteristic functions
	of $(X,Y)$, $X$ and $Y$, respectively,	 $w(u,v)= (c_p c_q \abs{u}_p^{1+p} \abs{v}_q^{1+q} )^{-1} $, 
	$c_d=\frac{\pi^{(1+d)/2}}{\Gamma((1+d)/2)}$ is half the surface area of the unit sphere 
	in ${\mathbb R}^{d+1}$, and $|\cdot|_d$ stands for the Euclidean norm in ${\mathbb R}^d$. 
	
In terms of ${\cal V}^2(X,Y)$, the square of the distance correlation is
\begin{equation}
	{\cal R}^2(X,Y)=\left\{\begin{array}{cl}
	\frac{{\cal V}^2(X,Y)}{\sqrt{{\cal V}^2(X,X){\cal V}^2(Y,Y)}},& {\cal V}^2(X){\cal V}^2(Y) > 0 \\ 
	0,& {\cal V}^2(X){\cal V}^2(Y) = 0.
	\end{array}\right.
\end{equation}
The distance correlation is a measure of statistical independence; that is,
 $\mathcal{R}^2(X,Y) = 0$ if and only if $X$ and $Y$ are independent. 
Besides being defined for random variables of different dimensions, it
has other valuable properties. In particular, it is rotationally invariant 
and scale equivariant \citep{sze12}.
A further advantage over other measures of independence, such as the 
mutual information, is that the distance correlation can be readily estimated 
using a plug-in estimator that does not involve any parameter tuning. 
The almost sure convergence of the estimator ${\cal V}_n^2$ is proved in \citet[Thm. 2]{sze07}. 
% Implementations of functions to estimate $\mathcal{V}_n^2$ and $\mathcal{R}_n^2$ 
% can be found in the R-package \textit{energy} by Sz\'ekely and Rizzo.
%[ASG: Find proper way of citation for the energy package
%JLT: Se podría citar el artículo "Energy statistics: A class of statistics based on distances", aunque también se podría eliminar la frase, la verdad es que es una implementación bastante mala (muy lenta) y es fácil de implementar, lo puse por completitud pero como vamos mal de espacio... 
% Por si acaso he añadido esta referencia y la anterior al final del references.bib]  

To summarize, in maxima hunting, one selects the $d$ different local maxima  
of the distance correlation between 
$X(t)$, the values of random process at different instants $t \in [0,1]$, and 
the response variable
\begin{equation} \label{eq:maxima_distance_correlation}
X(t_i) = \argmaxl_{t \in [0,1]} \mathcal{R}^2(X(t),Y), \quad i = 1,2, \ldots, d.
\end{equation} 
Maxima Hunting is easy to interpret. 
It is also well-motivated from the point of view of FDA, because it takes 
advantage of functional properties 
of the data, such as continuity, which implies that similar information is conveyed 
by the values of the function at neighboring points.
In spite of the simplicity of the method, it naturally accounts for the 
relevance and redundancy trade-off in feature selection \citep{yu04}: 
the local maxima (\ref{eq:maxima_distance_correlation}) are relevant for discrimination.
Points around them, which do not maximize the distance correlation with the class label, 
are automatically excluded. 
Furthermore, it is also possible to derive a uniform convergence result, which provides
additional theoretical support for the method.
Finally, the empirical investigation carried out in \cite{ber16mh} shows that MH 
performs well in standard benchmark classification problems for functional data.
In fact, for some problems, one can show that the optimal (Bayes) 
classification rules depends only on the maxima of $\mathcal{R}^2(X(t),Y)$. 

However, maxima hunting presents also some limitations. First, it is not always 
a simple task to estimate the local maxima, especially in functions that are very smooth or that
vary abruptly. Furthermore, there is no guarantee that different maxima are not redundant. 
In most cases, the local maxima of $\mathcal{R}^2(X(t),Y)$ 
are indeed relevant for classification. 
However, there are important points for which this quantity does not attain a maximum. 

As an example, consider the family of classification problems introduced in \citet[Prop. 3]{ber16mh},
in which the goal is to discriminate trajectories generated 
by a standard Brownian motion
process, $B(t)$, and trajectories from the process $B(t) + \Phi_{m,k}(t)$, where  
\begin{equation}
\Phi_{m,k}(t)=\int_0^t\sqrt{2^{m-1}}
\left[ \mathbb{I}_{\left( \frac{2k-2}{2^m},\frac{2k-1}{2^m}\right)}(s)  - 
\mathbb{I}_{\left( \frac{2k-1}{2^m},\frac{2k}{2^m}\right)}(s)\right]ds, \quad m,k\in{\mathbb N}, 1\leq k\leq 2^{m-1}. 
\end{equation}
Assuming a balanced class distribution ($\mathbb{P}(Y=0) = \mathbb{P}(Y=1) = 1/2$), 
the optimal classification rule is $g^*(x)=1  $ if and only if 
$
\left( X\left(\frac{2k-1}{2^m}\right) - X\left(\frac{2k-2}{2^m}\right)\right) +
\left( X\left(\frac{2k-1}{2^m}\right) - X\left(\frac{2k}{2^m}\right)\right) 
>
\frac{1}{\sqrt{2^{m+1}}}. %- \frac{1}{\sqrt{2^{m-1}}} \log\left(\frac{p}{1-p}\right).
$
The optimal classification error is
$
L^*= 1-\text{normcdf}\left(\frac{\parallel \Phi_{m,k}^\prime(t)\parallel}{2}\right)=1-\text{normcdf}\left(\frac{1}{2}\right)\simeq0.3085,
$
where, $\parallel \cdot \parallel$ denotes the $L^2[0,1]$ norm, 
and $\text{normcdf}(\cdot)$ is the cumulative distribution function of the standard normal.
The relevance function has a single maximum at $X\left(\frac{2k-1}{2^m}\right)$. 
However, the Bayes classification rule involves three relevant variables, two 
of which are clearly not maxima of $\mathcal{R}^2(X(t),Y)$.
In spite of the simplicity of these types of functional classification problems, they
are important to analyze, because the set of functions $\Phi_{m,k}$, 
with $m > 0$ and $k > 0$ form an orthonormal 
basis of the Dirichlet space  $\mathcal{D}[0,1]$, the space of continuous functions 
whose derivatives are in $L^2[0,1]$. Furthermore, this space is the reproducing kernel 
Hilbert space associated with Brownian motion and plays and important role in 
functional classification \citep{mor10,ber16rkhs}.
In fact, any trend in the Brownian process
can be approximated by a linear combination or by a mixture of $\Phi_{m,k}(t)$.  

To illustrate the workings of maxima hunting and its limitations  
we analyze in detail the classification problem 
$B(t)$ vs.  $B(t) + 2 \Phi_{3,3}(t)$, which is of the type considered above.
In this case, the optimal classification rule depends on the maximum $X(5/8)$, 
and on $X(1/2)$ and $X(3/4)$,  which are not maxima, and
would therefore not be selected by the MH algorithm.
 The optimal error is $L^*=15.87\%$. 
To illustrate the importance of selecting all the relevant variables, we  perform simulations 
in which we compare the accuracy of the linear Fisher discriminant with 
the maxima hunting selection, and with the optimal variable selection procedures. 
In these experiments, independent training and test samples 
of size $1000$ are generated.  
The values reported  are averages over $100$ independent runs.
Standard deviations are given between parentheses.
The average
prediction error when only the maximum of the trajectories is considered is $37.63\%(1.44\%)$. 
When all three variables are used the empirical error is $15.98\%(1\%)$, which is close to the Bayes error. 
When other points in addition to the maximum are used (i.e., $(X(t_1),X(5/8),X(t_2)$, with $t_1$ and $t_2$ 
randomly chosen so that $0\leq t_1<5/8<t_2\leq1$) the average classification error is $22.32\%(2.18\%)$. 
In the top leftmost plot of Figure \ref{fig:pico1} trajectories from both classes, 
together with the corresponding averages (thick lines) are shown.
The relevance function $\mathcal{R}^2(X(t),Y)$ is plotted below. 
The relevant variables, which are required for optimal classification,
 are marked by dashed vertical lines.  

\section{Recursive Maxima Hunting}\label{sec:mhrec} 

\begin{figure}
	\centering
	\includegraphics[width=0.95\linewidth,height=5cm]{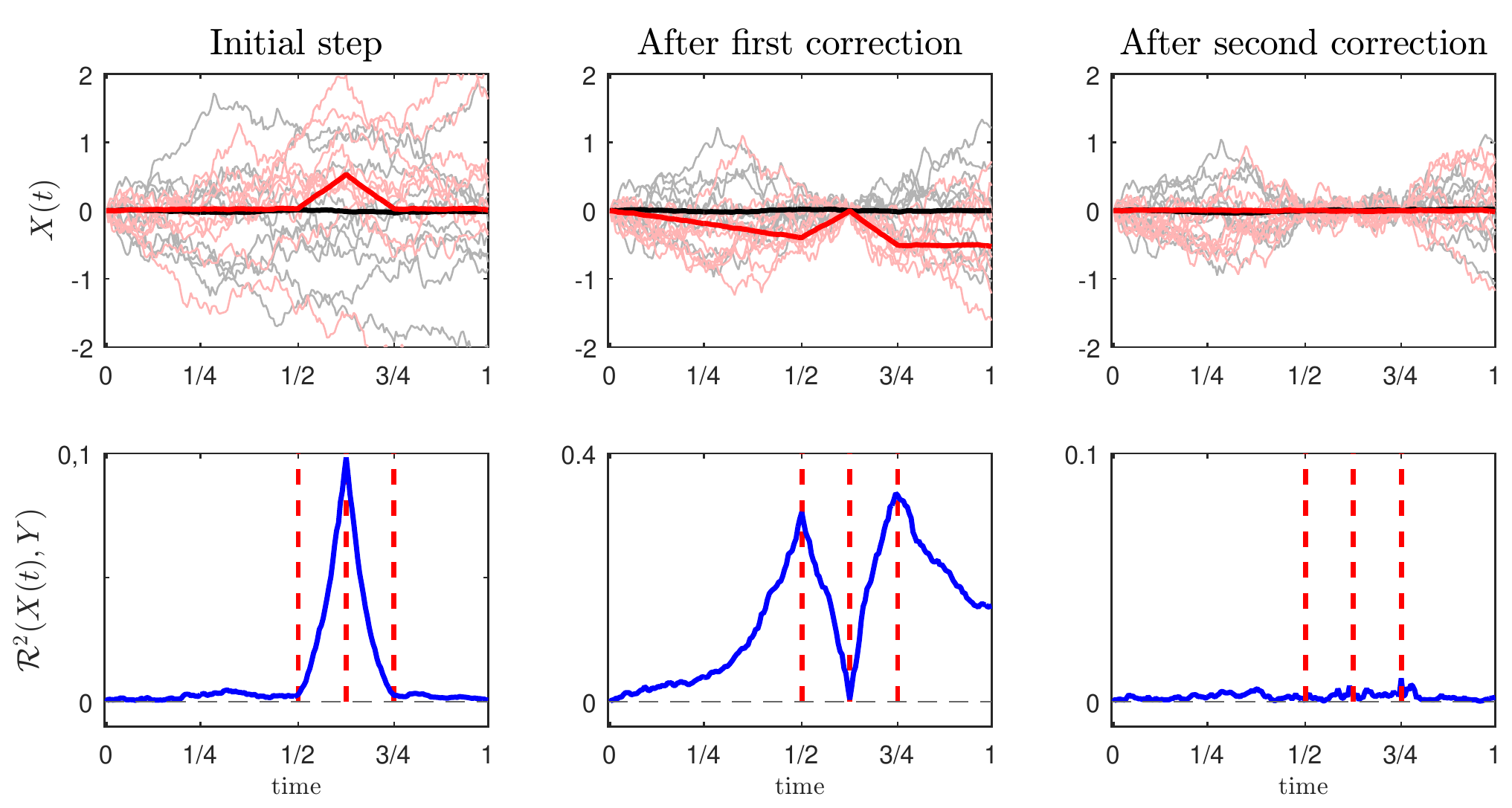}
	\caption{\footnotesize {\it First row}: Individual and average trajectories for the classification of $B(t)$ vs.  $B(t) + 2 \Phi_{3,3}(t)$ initially (left) and after the first (center) and second (right) corrections. {\it Second row}: Values of $\mathcal{R}^2(X(t),Y)$ as a function of t. The variables required for optimal classification are marked with vertical dashed lines.}
	\label{fig:pico1}
\end{figure}

As a variable selection process,  MH avoids, at least partially, 
the redundancy introduced by the continuity of the functions that characterize the instances. 
However, this local approach cannot detect redundancies among different 
local maxima. Furthermore, there could be points in the trajectory that do not
correspond to maxima of the relevance function, but which are relevant when 
considered jointly with the maxima. 
The goal of recursive maxima hunting (RMH) is to select the maxima of 
$\mathcal{R}^2(X(t),Y)$  in a recursive manner by removing at each step 
the information associated to the most recently selected maximum. 
This avoids the influence of previously selected maxima, which can obscure 
ulterior dependencies. The influence of a selected variable $X(t_0)$ on the rest 
of the trajectory can be eliminated by subtracting the conditional expectation 
$\mathbb{E}(X(t)|X(t_0))$ from $X(t)$. Assuming that the underlying process is Brownian 
\begin{equation}\label{eq:cond_exp_b}
\mathbb{E}(X(t)|X(t_0)) = \frac{\min(t,t_0)}{t_0} X(t_0), \quad t \in [0,1].
\end{equation}
In the subsequent iterations, there are two intervals:  $[t,t_0]$ and $[t_0,1]$. 
Conditioned on the value at $X(t_0)$, the process in the interval $[t_0,1]$ is 
still Brownian motion. By contrast, for the interval $[0,t_0]$ 
the process is a Brownian bridge, whose conditional expectation is
\begin{equation}\label{eq:cond_exp_bb}
\mathbb{E}(X(t)|X(t_0)=\frac{\min(t,t_0)-t \, t_0}{ t_0 (1- t_0)}X(t_0)=
\left \{ 
\begin{matrix}
\frac{t}{t_0}X(t_0),  & t < t_0\\
\frac{1-t}{1-t_0}X(t_0), & t > t_0.
\end{matrix}\right.
\end{equation}
As illustrated by the results in the experimental section, the Brownian hypothesis 
is a robust assumption. Nevertheless, if additional information on the 
underlying stochastic processes is available, it can be incorporated to 
the algorithm during the calculation of the conditional expectation 
in Equations (\ref{eq:cond_exp_b}) and (\ref{eq:cond_exp_bb}). %\textcolor{red}{fórmula general para Gaussianos?}

%\begin{figure}
%	\centering
%	\includegraphics[width=0.75\linewidth,height=4cm]{Figures/pico2}
%	\caption{\footnotesize Left: trajectories and mean functions of the toy example after the first correction. Right: associated distance correlation function $\mathcal{R}^2(X(t),Y)$.}
%	\label{fig:pico2}
%\end{figure}
The center and right plots in Figure \ref{fig:pico1} illustrate the behavior of RMH in the example described in the previous section. The top center plot diplays the trajectories 
and corresponding averages (thick lines)  for both classes after applying the correction 
(\ref{eq:cond_exp_b}) with $t_0=5/8$, which is the first maximum of the distance correlation function (bottom leftmost plot in Figure \ref{fig:pico1}). 
The variable $X(5/8)$ is clearly uninformative once this correction has been applied.
The distance correlation $\mathcal{R}^2(X(t),Y)$  for the corrected trajectories 
is displayed in the bottom center plot.
Also in this plot the relevant variables are marked by vertical dashed lines. 
It is clear that the subsequent local maxima at $ t = 1/2$, in the subinterval $ [0, 5/8]$,
and at $t =3/4$, in the subinterval, and $ [5/8, 1]$ 
correspond to the remaining relevant variables.
The last column shows the corresponding plots after the correction is applied anew
 (equations (\ref{eq:cond_exp_bb}) with $t_0=1/2$ in $ [0, 5/8]$ and  (\ref{eq:cond_exp_b}) with $t_0=3/4$ in $ [5/8,1]$). 
After this second correction, the discriminant information has been removed. 
In consequence, the distance correlation function, up to sample fluctuations, is zero. 

An important issue in the application of this method is how to decide when 
to stop the recursive search. The goal is to avoid including irrelevant 
and/or redundant variables. 
To address the first problem, we only include maxima that are sufficiently prominent
$\mathcal{R}^2(X(t_{max}),Y) > s$, where $0 < s < 1$ can be used to gauge the 
relative importance of the maximum.
Redundancy is avoided by excluding points around a selected maximum $t_{max}$ 
for which  $\mathcal{R}^2(X(t_{max}),X(t))\geq r$, for some redundancy threshold 
$ 0 < r <1$, which is typically close to one. As a result of these two conditions 
only a finite (typically small) number of variables are selected. 
This data-driven stopping criterion avoids the need to set the number of selected variables 
beforehand or to determine this number by a costly validation procedure. 
The sensitivity of the results to the values of $r$ and $s$ will be studied in Section \ref{sec:exp}.
Nonetheless, RMH has a good and robust performance for a wide range of
reasonable values of these parameters ($r$ close to $1$ and $s$ close to 0). 
The pseudocode of the RMH algorithm is given in Algorithm 1.

% \vspace{-0.2cm}%
\alglanguage{pseudocode}
\begin{algorithm}[!htp]\label{ddalgorithm}
	\small
	\caption{Recursive Maxima Hunting}
	\label{Algorithm:delete_mCBF_aCBF}
	\begin{algorithmic}[1]
		\Function{\textbf{RMH}}{$X(t), Y$}
		\State $\mathbf{t}^* \gets \left[ \ \right]$		\Comment Vector of selected points initially empty
		\State RMH\_rec($X(t)$,$Y$,$0$,$1$)             \Comment Recursive search of the maxima of $\mathcal{R}^2(X(t),Y)$ 
		\State \Return $\mathbf{t}^*$                   \Comment Vector of selected points
		\EndFunction 
		\Procedure{\textbf{RMH\_rec}}{$X(t), Y , t_{inf}, t_{sup}$}
		\State{$t_{max} \gets \argmaxl_{t_{inf} \le t \le t_{sup}} \left\{ \mathcal{R}^2(X(t),Y) \right\}$}	
		\If {$\mathcal{R}^2(X(t_{max}),Y) > s$}
		\State $\mathbf{t}^* \gets \left[\mathbf{t}^* \ t_{max} \right]$ \Comment Include $t_{max}$ in $\mathbf{t}^*$ the vector of selected points
		\State{$X(t) \gets X(t) - \mathbb{E}(X(t)\mid X(t_{max})), \ t \in [t_{inf},t_{sup}] $} 
		\Comment Correction of type (\ref{eq:cond_exp_b}) or (\ref{eq:cond_exp_bb}) as required
		\Else
           \State \Return
		\EndIf
		\State $\triangleright$ Exclude redundant points	to the left of $t_{max}$	
		\State $t_{max}^{-}  \gets \max\limits_{ t_{inf} \le t < t_{max}}  \left\{ t :   
		                        \mathcal{R}^2\left(X(t_{max}), X(t)\right) \le r \right\} $  
		\If{$t_{max}^{-} > t_{inf}$} 
			\State RMH\_rec($X(t), Y, t_{inf}, t_{max}^{-}$)  \Comment Recursion on left subinterval
		\EndIf
		\State $\triangleright$ Exclude redundant points to the right of $t_{max}$		
		\State $t_{max}^{+}  \gets \min\limits_{ t_{max} < t \le t_{sup}} \left\{ t :   
		                        \mathcal{R}^2\left(X(t_{max}), X(t) \right) \le r\right\} $  
		\If{$t_{max}^{+} < t_{sup}$} 
			\State RMH\_rec($X(t), Y, t_{max}^{+}, t_{sup}$)  \Comment Recursion on right subinterval
		\EndIf
		\State \Return		
		\EndProcedure
		\Statex
	\end{algorithmic}
	\vspace{-0.4cm}%
\end{algorithm}

%\textcolor{red}{¿Algo de teoría? En teoría se mantienen los resultados de \citep{ber16mh}, ¿pensar algo mas?...}

\section{Empirical study}\label{sec:exp}

To assess the performance of RMH, we have carried out experiments in 
simulated and real-world data in which it is compared with some 
well-established dimensionality reduction methods, such as PCA \citep{ram05} and
partial least squares \citep{del12pls}, and with Maxima Hunting
\citep{ber16mh}. In these experiments, $k$-nearest neighbors 
 (kNN) with the Euclidean distance is used for classification. 
kNN has been selected because it is a simple, nonparametric classifier 
with reasonable overall predictive accuracy.
The value $k$ in kNN is selected by 10-fold CV from integers in $[1,\sqrt{N_{train}}]$, 
where $N_{train}$ is the size of the training set.
Since RMH is a filter method for variable selection, the results are 
expected to be similar when other types of classifiers are used.
As a reference, the results of kNN using complete trajectories 
(i.e., without dimensionality reduction) are also reported. 
This approach is referred to as \textit{Base}. 
Note that, in this case, the performance of kNN need not be optimal 
because of the presence of irrelevant attributes. 

RMH requires determining the values of two hyperparameters: 
the redundancy threshold  $r$ ($ 0 < r < 1$ typically close to $1$),
and the relevance threshold $s$   ($ 0 < s < 1$ typically close to $0$). 
Through extensive simulations we have observed that 
RMH is quite robust for a wide range of appropriate values of these parameters. 
In particular, the results are very similar for values of r in the interval 
$[0.75, 0.95]$. 
The predictive accuracy is somewhat more sensitive to the choice of $s$: 
If the value of $s$ is too small, irrelevant variables can be selected.
If $s$ is too large, it is possible that relevant points are excluded. 
For most of the experiments performed, the optimal values of $s$ are between $0.025$ and $0.1$. 
In view of these observations, the experiments are made using $r=0.8$. 
The value of $s$ is selected from the set $\{0.025,0.05,0.1\}$ by 10-fold CV. 
A more careful determination of $r$ and $s$ is beneficial, especially in some 
extreme problems (e.g., with very smooth or with rapidly-varying trajectories).
In RMH, the number of selected variables, which is not determined beforehand, 
depends indirectly on the values of $r$ and $s$. 
In the other methods, the number of selected variables 
is determined using 10-fold CV, with maximum of $30$. 

\begin{figure}
	\centering
	\includegraphics[width=0.9\linewidth,height=3.5cm]{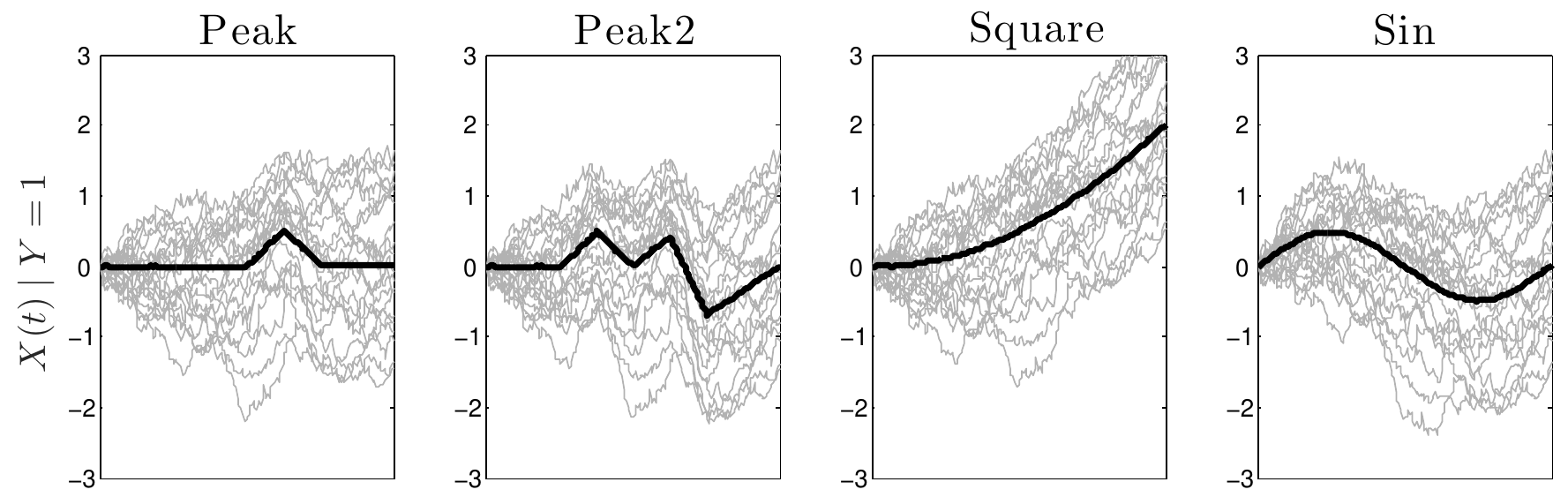}
	\caption{\footnotesize Class $1$ trajectories and averages (thick lines) for the different synthetic problems.}
	\label{fig:sim}
\end{figure}

A first batch of experiments is carried out on simulated data generated from the model
$$\left\{\begin{array}{lcl}
P_0: B(t)&,& t\in[0,1] \\ 
P_1: B(t)+m(t)&,& t\in[0,1]
\end{array}\right. ,$$
where $B(t)$ is standard Brownian motion, $m(t)$ is a deterministic trend, 
and $\mathbb{P}(Y=0)=\mathbb{P}(Y=1)=1/2$. 
Using \citet[Theorem 2]{ber16rkhs}, it is possible to compute the optimal 
classification rules $g^*$ and the corresponding Bayes errors $L^*$. 
To ensure a wide coverage, we consider two problems in which the Bayes rule 
depends only on a few variables and two problems in which complete 
trajectories are needed for optimal classification: 
(i) \textit{Peak}: $m(t)=2\Phi_{3,3}(t)$. The optimal rule depends only 
on $X(1/2)$, $X(5/8)$ and $X(3/4)$. The Bayes error is $L^*\simeq 0.1587$.  
This is the example analyzed in the previous section. 
(ii)  \textit{Peak2}: $m(t)=2\Phi_{3,2}(t)+3\Phi_{3,3}(t)-2\Phi_{2,2}(t)$.  
The optimal rule depends only on $X(1/4), X(3/8), X(1/2),$ $X(5/8), X(3/4)$, 
and $X(1)$. The Bayes error is $L^*\simeq 0.0196$.
(iii) \textit{Square}: $m(t) = 2t^2$. The Bayes error is $L^*\simeq 0.1241$.
(iv)  \textit{Sin}: $m(t) = 1/2\sin(2\phi t)$.  The Bayes error is $L^*\simeq 0.1333$.
In Figure \ref{fig:sim} we have plotted some trajectories corresponding
 to class $1$ instances, together with their corresponding averages (thick lines). 
Class $0$ trajectories are realizations of a standard Brownian process. 
In these experiments, training samples of different sizes ($N_{train}=\{50, 100, 200, 500, 1000\}$) 
and an independent test set of size $1000$ are generated. 
The trajectories are discretized in $200$ points. 
Half of the trajectories belong to each class in both the training and test sets. 
The values reported are averages over $200$ independent repetitions.

\begin{figure}
	\centering
	\includegraphics[width=0.9\linewidth]{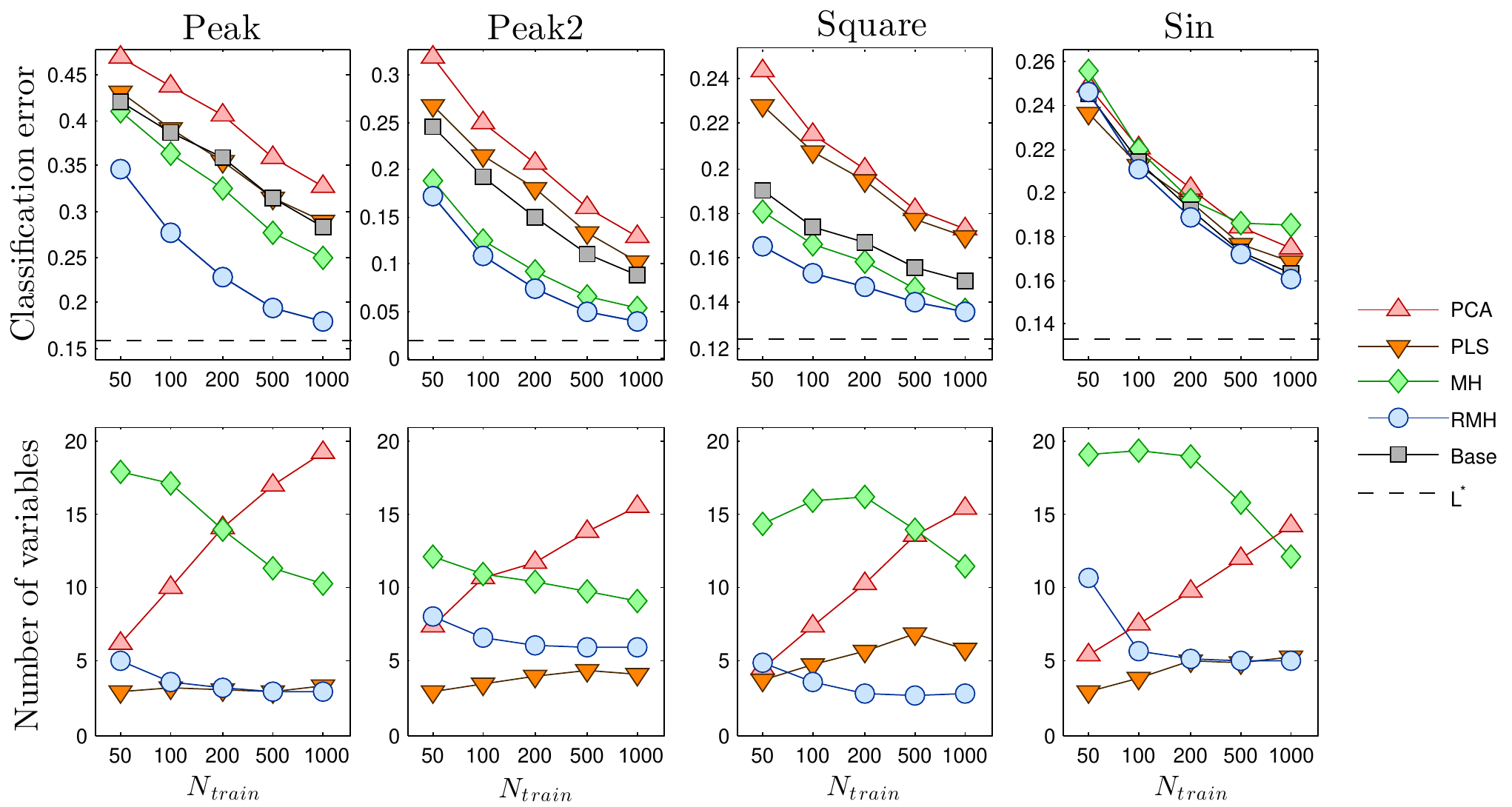}
	\caption{\footnotesize Average classification error (first row) and average number of selected variables/components (second row) as a function of the size of the training. 
}
	\label{fig:err_sim}
\end{figure}

Figure \ref{fig:err_sim} displays the average classification error (first row) 
and the average number of selected variable
/components (second row) as a function of the training sample size 
for each model and classification method. 
Horizontal dashed lines are used to indicate the Bayes error level in the different problems.   
From the results reported in Figure \ref{fig:err_sim}, one concludes that RMH 
has the best overall performance.
It is always more accurate than the \textit{Base} method. 
This observation justifies performing variable selection not only for the sake 
of dimensionality reduction, but also to improve the classification accuracy. 
RMH is also better than the original MH in all the problems investigated: 
there is both an improvement of the prediction error, and  a reduction of the 
numeber of variables used for classification.
In \textit{peak} and \textit{peak2}, problems in which the relevant 
variables are known, RMH generally selects the correct ones. 
As expected, PLS performs better than PCA. 
However, both MH and RMH outperform these projection methods, 
except in \textit{sin}, where their accuracies are similar. 
Both PLS and RMH are effective dimensionality reduction methods with comparable performance.
However, the components selected in PLS are, in general, more difficult to interpret
because they involve whole trajectories.
Finally, the accuracy of RMH is very close to the Bayes level for higher sample sizes, 
even when the optimal rule requires using complete trajectories 
(\textit{square} and \textit{sin}). 

\begin{figure}
	\centering
	\includegraphics[width=0.9\linewidth,height=4.8cm]{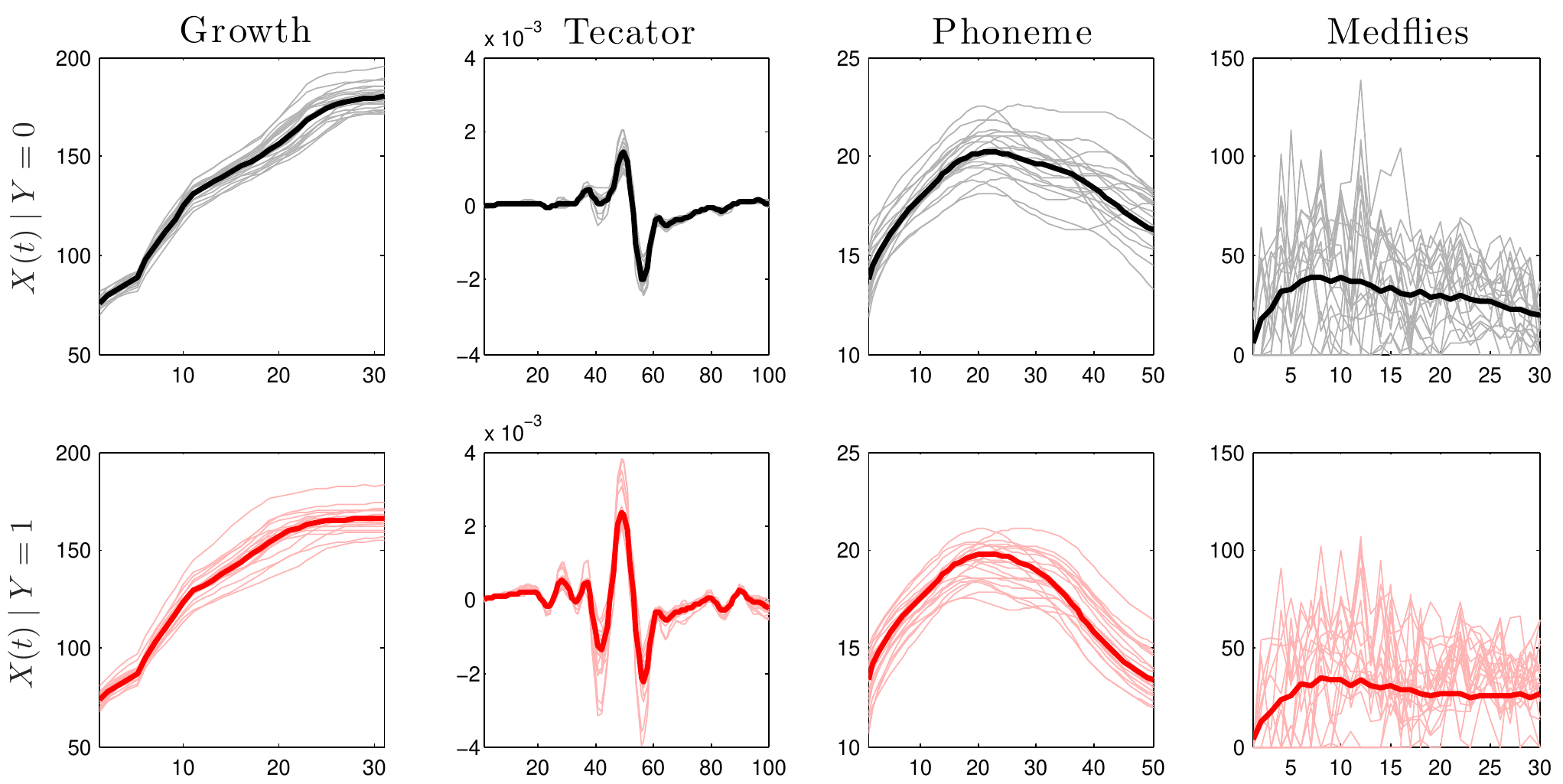}
	\caption{\footnotesize Trajectories for each of the classes and their corresponding averages (thick lines).}
	\label{fig:real}
\end{figure}

\begin{figure}
	\centering
	\includegraphics[width=0.85\linewidth]{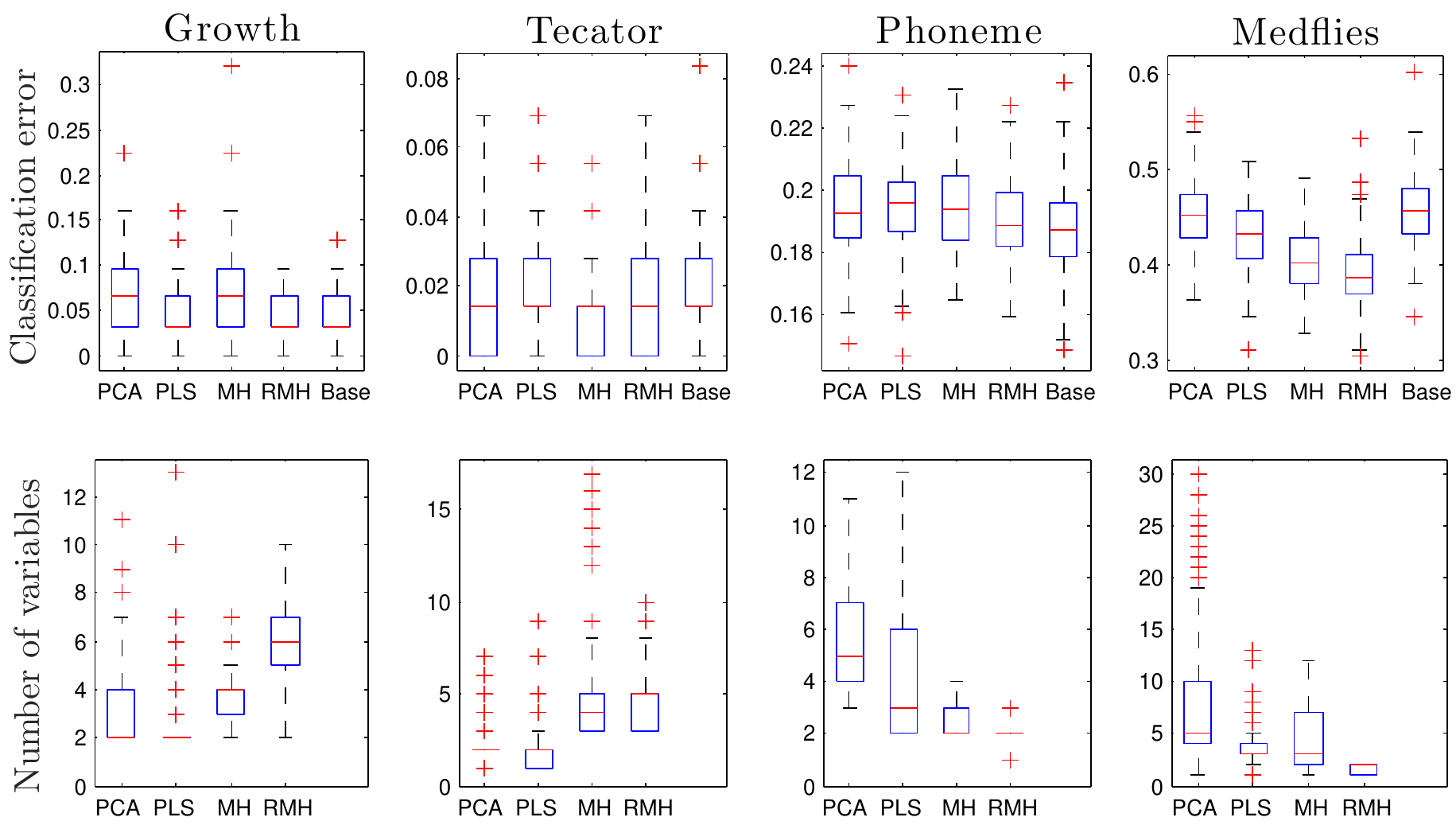}
	\caption{\footnotesize Classification error (first row) and number of variables/components selected (second row) by RMH.}
	\label{fig:err_real}
\end{figure}

To assess the performance of RMH in real-world functional classification problems, 
we have carried out a second batch of experiments in four datasets, which are commonly 
used as benchmarks in the FDA literature. Instances in \textit{Growth} correspond to curves of the heights of 54 girls and 38 boys from the \textit{Berkeley Growth Study}. 
Observations are discretized in $31$ non-equidistant ages between $1$ and $18$ years \citep{ram05,mos14}. The \textit{Tecator} dataset consists of $215$ near-infrared absorbance spectra of finely chopped meat. The spectral curves consist of $100$ equally spaced points. The class labels are determined in terms of fat content (above or below $20\%$). 
The curves are fairly smooth. In consequence, we have followed the general recommendation and used the second derivative for classification \citep{fer06,gal14}.
The \textit{Phoneme} data consists of $4509$ log-periodograms observed 
at $256$ equidistant points. Here, we consider the binary problem of 
distinguishing between the phonemes ``aa'' ($695$) and ``ao'' ($1022$) \citep{gal14}. 
Following \cite{del12np}, the curves are smoothed with a local linear method 
and truncated to the first $50$ variables.
The \textit{Medflies} are records of daily egg-laying patterns of a thousand flies. 
The goal is to discriminate between short- and long-lived flies. 
Following \cite{mos14}, curves equal to zero are excluded. 
There are $512$ $30$-day curves (starting from day $5$) of flies who 
live at most $34$ days, $266$ of these are long-lived (reach the day $44$). 
The classes in \textit{Growth} and \textit{Tecator} are well separated. 
In consequence, they are relatively easy problems. 
By contrast, \textit{Phoneme } and \textit{Medflies} are notoriously 
difficult classification tasks.  
Some trajectories of each problem and each class, together with the corresponding 
averages (thick lines), are plotted in Figure \ref{fig:real}.
To estimate the classification error, the datasets are partitioned at random into a training set (with $2/3$ of the observations) and a test set ($1/3$). This procedure is repeated $200$ times. The boxplots of the results for each dataset and method are shown in Figure \ref{fig:err_real}. Errors are shown in first row and the number of selected variables/components in the second one.
From these results we observe that, in general, dimensionality reduction is effective: 
the accuracy of the four considered methods is similar or better than 
the \textit{Base} method, in which complete trajectories are used for classification. 
In particular, \textit{Base} does not perform well when the trajectories 
are not smooth (\textit{Medflies}).
The best overall performance corresponds to RMH. 
In the easy problems (\textit{Growth} and \textit{Tecator}), 
all methods behave similarly and give good results.
In \textit{Growth}, RMH is slightly more accurate. 
However, it tends to select more variables than the other methods. 
In the more difficult problems, (\textit{Phoneme} and \textit{Medflies}),
RMH yields very accurate predictions while selecting only two variables. 
In these problems it exhibits the best performance,  except in 
\textit{Phoneme}, where \textit{Base} is more accurate. 
The variables selected by RMH and MH are directly interpretable, 
which is an advantage over projection-based methods (PCA, PLS).
Finally, let us point out that the accuracy of RMH is comparable and often
better that state-of-the-art functional classification methods.
See, for instance, \cite{ber16mrmr,del12cw,del12np,mos14,gal14}. 
In most of these works no dimensionality reduction is applied.
Nevertheless, these comparisons must be done carefully because the evaluation 
protocol and the classifiers used vary in the different studies. 
In any case, RMH is a filter method, which means that it could be more effective 
if used in combination with other types of classifiers or adapted and 
used as a wrapper or, even, as an embedded variable selection method.

% \section{Conclusions}\label{sec:conclusions}

%  We have proposed a new filter variable selection method, recursive maxima hunting. 
% MHR provides an interpretable reduction of  functional data  when the final goal is the supervised classification.
% The iterative and recursive procedure allow us to consider the interactions between 
% variables overcoming the main drawbacks of the original maxima hunting but keeping its advantages. 
% From the empirical point of view, RMH clearly outperforms MH and the state-of-art projection methods (PLS and PCA). It also performs better than the direct classification of the entire curves, what is another reason in favour of the dimension reduction before functional classification.

% RMH give us a reduced set of variables which can be used as an estimator of the number of relevant variables (a difficult open problem). This subset of variables is quite accurate in many cases and independent of the learning model. Nevertheless, some times too many variables are chosen (\textit{Tecator} and \textit{Growth}) what makes us think the  method can be refined with further research.

\subsubsection*{Acknowledgments}
The authors thank Dr. Jos\'e R. Berrendero for his insightful suggestions.
We also acknowledge financial support from the Spanish Ministry of Economy and Competitiveness, project TIN2013-42351-P and from the Regional Government of Madrid, CASI-CAM-CM project (S2013/ICE-2845).

\setlength{\bibsep}{0.0pt}\small

\newpage

\bibliography{references}{}

\begin{thebibliography}{}

\bibitem[\protect\citeauthoryear{Aneiros and Vieu}{Aneiros and
  Vieu}{2014}]{ane14}
Aneiros, G. and P.~Vieu (2014).
\newblock Variable selection in infinite-dimensional problems.
\newblock {\em Statistics \& Probability Letters\/}~{\em 94}, 12--20.

\bibitem[\protect\citeauthoryear{Ba\'{\i}llo, Cuevas, and Fraiman}{Ba\'{\i}llo
  et~al.}{2011}]{bai11}
Ba\'{\i}llo, A., A.~Cuevas, and R.~Fraiman (2011).
\newblock {\em Classification methods for functional data}, pp.\  259--297.
\newblock Oxford: Oxford University Press.

\bibitem[\protect\citeauthoryear{Berrendero, Cuevas, and Torrecilla}{Berrendero
  et~al.}{2015}]{ber16rkhs}
Berrendero, J.~R., A.~Cuevas, and J.~L. Torrecilla (2015).
\newblock {On near perfect classification and functional Fisher rules via
  reproducing kernels}.
\newblock {\em arXiv:1507.04398\/}, 1--27.

\bibitem[\protect\citeauthoryear{Berrendero, Cuevas, and Torrecilla}{Berrendero
  et~al.}{2016a}]{ber16mrmr}
Berrendero, J.~R., A.~Cuevas, and J.~L. Torrecilla (2016a).
\newblock {The mRMR variable selection method: a comparative study for
  functional data}.
\newblock {\em Journal of Statistical Computation and Simulation\/}~{\em
  86\/}(5), 891--907.

\bibitem[\protect\citeauthoryear{Berrendero, Cuevas, and Torrecilla}{Berrendero
  et~al.}{2016b}]{ber16mh}
Berrendero, J.~R., A.~Cuevas, and J.~L. Torrecilla (2016b).
\newblock Variable selection in functional data classification: a maxima
  hunting proposal.
\newblock {\em Statistica Sinica\/}~{\em 26\/}(2), 619--638.

\bibitem[\protect\citeauthoryear{Delaigle and Hall}{Delaigle and
  Hall}{2012a}]{del12np}
Delaigle, A. and P.~Hall (2012a).
\newblock Achieving near perfect classification for functional data.
\newblock {\em Journal of the Royal Statistical Society B\/}~{\em 74\/}(2),
  267--286.

\bibitem[\protect\citeauthoryear{Delaigle and Hall}{Delaigle and
  Hall}{2012b}]{del12pls}
Delaigle, A. and P.~Hall (2012b).
\newblock Methodology and theory for partial least squares applied to
  functional data.
\newblock {\em The Annals of Statistics\/}~{\em 40\/}(1), 322--352.

\bibitem[\protect\citeauthoryear{Delaigle, Hall, and Bathia}{Delaigle
  et~al.}{2012}]{del12cw}
Delaigle, A., P.~Hall, and N.~Bathia (2012).
\newblock Componentwise classification and clustering of functional data.
\newblock {\em Biometrika\/}~{\em 99\/}(2), 299--313.

\bibitem[\protect\citeauthoryear{Ding and Peng}{Ding and Peng}{2005}]{din05}
Ding, C. and H.~Peng (2005).
\newblock Minimum redundancy feature selection from microarray gene expression
  data.
\newblock {\em Journal of Bioinformatics and Computational Biology\/}~{\em
  3\/}(2), 185--205.

\bibitem[\protect\citeauthoryear{Fernandez-Lozano, Seoane, Gestal, Gaunt,
  Dorado, and Campbell}{Fernandez-Lozano et~al.}{2015}]{fer15}
Fernandez-Lozano, C., J.~A. Seoane, M.~Gestal, T.~R. Gaunt, J.~Dorado, and
  C.~Campbell (2015).
\newblock Texture classification using feature selection and kernel-based
  techniques.
\newblock {\em Soft Computing\/}~{\em 19\/}(9), 2469--2480.

\bibitem[\protect\citeauthoryear{Ferraty, Hall, and Vieu}{Ferraty
  et~al.}{2010}]{fer10}
Ferraty, F., P.~Hall, and P.~Vieu (2010).
\newblock Most-predictive design points for functional data predictors.
\newblock {\em Biometrika\/}~{\em 97\/}(4), 807--824.

\bibitem[\protect\citeauthoryear{Ferraty and Vieu}{Ferraty and
  Vieu}{2006}]{fer06}
Ferraty, F. and P.~Vieu (2006).
\newblock {\em {Nonparametric Functional Data Analysis: Theory and Practice}}.
\newblock Springer.

\bibitem[\protect\citeauthoryear{Fraiman, Gim\'enez, and Svarc}{Fraiman
  et~al.}{2016}]{fra16}
Fraiman, R., Y.~Gim\'enez, and M.~Svarc (2016).
\newblock Feature selection for functional data.
\newblock {\em Journal of Multivariate Analysis\/}~{\em 146}, 191--208.

\bibitem[\protect\citeauthoryear{Galeano, Joseph, and Lillo}{Galeano
  et~al.}{2014}]{gal14}
Galeano, P., E.~Joseph, and R.~E. Lillo (2014).
\newblock {The Mahalanobis distance for functional data with applications to
  classification}.
\newblock {\em Technometrics\/}~{\em 57\/}(2), 281--291.

\bibitem[\protect\citeauthoryear{G\'{o}mez-Verdejo, Verleysen, and
  Fleury}{G\'{o}mez-Verdejo et~al.}{2009}]{gom09}
G\'{o}mez-Verdejo, V., M.~Verleysen, and J.~Fleury (2009).
\newblock Information-theoretic feature selection for functional data
  classification.
\newblock {\em Neurocomputing\/}~{\em 72\/}(16), 3580--3589.

\bibitem[\protect\citeauthoryear{Grosenick, Greer, and Knutson}{Grosenick
  et~al.}{2008}]{gro08}
Grosenick, L., S.~Greer, and B.~Knutson (2008).
\newblock {Interpretable classifiers for FMRI improve prediction of purchases}.
\newblock {\em Neural Systems and Rehabilitation Engineering, IEEE Transactions
  on\/}~{\em 16\/}(6), 539--548.

\bibitem[\protect\citeauthoryear{Guyon, Gunn, Nikravesh, and Zadeh}{Guyon
  et~al.}{2006}]{guy06}
Guyon, I., S.~Gunn, M.~Nikravesh, and L.~A. Zadeh (2006).
\newblock {\em {Feature Extraction: Foundations and Applications}}.
\newblock Springer.

\bibitem[\protect\citeauthoryear{Kneip and Sarda}{Kneip and
  Sarda}{2011}]{kne11}
Kneip, A. and P.~Sarda (2011).
\newblock Factor models and variable selection in high-dimensional regression
  analysis.
\newblock {\em The Annals of Statistics\/}~{\em 39\/}(5), 2410--2447.

\bibitem[\protect\citeauthoryear{Li and Yu}{Li and Yu}{2008}]{li08}
Li, B. and Q.~Yu (2008).
\newblock {Classification of functional data: A segmentation approach}.
\newblock {\em Computational Statistics \& Data Analysis\/}~{\em 52\/}(10),
  4790--4800.

\bibitem[\protect\citeauthoryear{Lindquist and McKeague}{Lindquist and
  McKeague}{2009}]{lin09}
Lindquist, M.~A. and I.~W. McKeague (2009).
\newblock {Logistic regression with Brownian-like predictors}.
\newblock {\em Journal of the American Statistical Association\/}~{\em
  104\/}(488), 1575--1585.

\bibitem[\protect\citeauthoryear{McKeague and Sen}{McKeague and
  Sen}{2010}]{mck10}
McKeague, I.~W. and B.~Sen (2010).
\newblock Fractals with point impact in functional linear regression.
\newblock {\em Annals of Statistics\/}~{\em 38\/}(4), 2559.

\bibitem[\protect\citeauthoryear{M{\"o}rters and Peres}{M{\"o}rters and
  Peres}{2010}]{mor10}
M{\"o}rters, P. and Y.~Peres (2010).
\newblock {\em Brownian Motion}.
\newblock Cambridge University Press.

\bibitem[\protect\citeauthoryear{Mosler and Mozharovskyi}{Mosler and
  Mozharovskyi}{2014}]{mos14}
Mosler, K. and P.~Mozharovskyi (2014).
\newblock {Fast DD-classification of functional data}.
\newblock {\em Statistical Papers\/}~{\em 55}, 49--59.

\bibitem[\protect\citeauthoryear{Preda, Saporta, and L\'{e}v\'{e}der}{Preda
  et~al.}{2007}]{pre07}
Preda, C., G.~Saporta, and C.~L\'{e}v\'{e}der (2007).
\newblock {PLS classification of functional data}.
\newblock {\em Computational Statistics\/}~{\em 22\/}(2), 223--235.

\bibitem[\protect\citeauthoryear{Ramsay and Silverman}{Ramsay and
  Silverman}{2005}]{ram05}
Ramsay, J.~O. and B.~W. Silverman (2005).
\newblock {\em {Functional Data Analysis}}.
\newblock Springer.

\bibitem[\protect\citeauthoryear{Ryali, Supekar, Abrams, and Menon}{Ryali
  et~al.}{2010}]{rya10}
Ryali, S., K.~Supekar, D.~A. Abrams, and V.~Menon (2010).
\newblock {Sparse logistic regression for whole-brain classification of fMRI
  data}.
\newblock {\em NeuroImage\/}~{\em 51\/}(2), 752--764.

\bibitem[\protect\citeauthoryear{Sz\'{e}kely and Rizzo}{Sz\'{e}kely and
  Rizzo}{2012}]{sze12}
Sz\'{e}kely, G.~J. and M.~L. Rizzo (2012).
\newblock On the uniqueness of distance covariance.
\newblock {\em Statistics \& Probability Letters\/}~{\em 82\/}(12), 2278--2282.

\bibitem[\protect\citeauthoryear{Sz\'{e}kely, Rizzo, and Bakirov}{Sz\'{e}kely
  et~al.}{2007}]{sze07}
Sz\'{e}kely, G.~J., M.~L. Rizzo, and N.~K. Bakirov (2007).
\newblock Measuring and testing dependence by correlation of distances.
\newblock {\em The Annals of Statistics\/}~{\em 35\/}(6), 2769--2794.

\bibitem[\protect\citeauthoryear{Tian and James}{Tian and James}{2013}]{tia13}
Tian, T.~S. and G.~M. James (2013).
\newblock Interpretable dimension reduction for classifying functional data.
\newblock {\em Computational Statistics \& Data Analysis\/}~{\em 57\/}(1),
  282--296.

\bibitem[\protect\citeauthoryear{Xiaobo, Jiewen, Povey, Holmes, and
  Hanpin}{Xiaobo et~al.}{2010}]{xia10}
Xiaobo, Z., Z.~Jiewen, M.~J. Povey, M.~Holmes, and M.~Hanpin (2010).
\newblock Variables selection methods in near-infrared spectroscopy.
\newblock {\em Analytica Chimica Acta\/}~{\em 667\/}(1), 14--32.

\bibitem[\protect\citeauthoryear{Yu and Liu}{Yu and Liu}{2004}]{yu04}
Yu, L. and H.~Liu (2004).
\newblock Efficient feature selection via analysis of relevance and redundancy.
\newblock {\em The Journal of Machine Learning Research\/}~{\em 5}, 1205--1224.

\bibitem[\protect\citeauthoryear{Zhou, Wang, and Wang}{Zhou
  et~al.}{2013}]{zho13}
Zhou, J., N.-Y. Wang, and N.~Wang (2013).
\newblock Functional linear model with zero-value coefficient function at
  sub-regions.
\newblock {\em Statistica Sinica\/}~{\em 23\/}(1), 25--50.

\end{thebibliography}

\end{document}